\documentclass[10pt,twocolumn,letterpaper]{article}

\usepackage{iccv}
\usepackage{times}
\usepackage{epsfig}
\usepackage{graphicx}
\usepackage{amsmath}
\usepackage{amssymb}
\usepackage{algorithm}
\usepackage{algorithmic}
\usepackage{booktabs}
\usepackage{multirow}
\usepackage{float}
\usepackage{subfigure}

\usepackage[pagebackref=true,breaklinks=true,letterpaper=true,colorlinks,bookmarks=false]{hyperref}

\iccvfinalcopy 

\def\iccvPaperID{3825} 

\ificcvfinal\fi

\begin{document}

\title{Voxel Transformer for 3D Object Detection}

\author{Jiageng Mao $^1$$^*$
\and
Yujing Xue $^2$$^*$
\and
Minzhe Niu $^3$
\and
Haoyue Bai $^4$
\and
Jiashi Feng $^2$
\and
Xiaodan Liang $^5$
\and
Hang Xu $^3$$^{\dag}$
\and
Chunjing Xu $^3$
}

\maketitle
\ificcvfinal\thispagestyle{empty}\fi

\begin{abstract}
We present Voxel Transformer (VoTr), a novel and effective voxel-based Transformer backbone for 3D object detection from point clouds. Conventional 3D convolutional backbones in voxel-based 3D detectors cannot efficiently capture large context information, which is crucial for object recognition and localization, owing to the limited receptive fields. In this paper, we resolve the problem by introducing a Transformer-based architecture that enables long-range relationships between voxels by self-attention. Given the fact that non-empty voxels are naturally sparse but numerous, directly applying standard Transformer on voxels is non-trivial. To this end, we propose the \textit{sparse voxel module} and the \textit{submanifold voxel module}, which can operate on the empty and non-empty voxel positions effectively. To further enlarge the attention range while maintaining comparable computational overhead to the convolutional counterparts, we propose two attention mechanisms for multi-head attention in those two modules: \textit{Local Attention} and \textit{Dilated Attention}, and we further propose \textit{Fast Voxel Query} to accelerate the querying process in multi-head attention. VoTr contains a series of sparse and submanifold voxel modules, and can be applied in most voxel-based detectors. Our proposed VoTr shows consistent improvement over the convolutional baselines while maintaining computational efficiency on the KITTI dataset and the Waymo Open dataset.
\end{abstract}

\let\thefootnote\relax\footnotetext{$^*$ Equal contribution. $^1$ The Chinese University of Hong Kong $^2$ National University of Singapore $^3$ Huawei Noah's Ark Lab $^4$ HKUST $^5$ Sun Yat-Sen University $^{\dag}$ Corresponding author: \url{xu.hang@huawei.com}}

\section{Introduction}
3D object detection has received increasing attention in autonomous driving and robotics. Detecting 3D objects from point clouds remains challenging to the research community, mainly because point clouds are naturally sparse and unstructured. Voxel-based detectors transform irregular point clouds into regular voxel-grids and show superior performance in this task. In this paper, we propose Voxel Transformer (VoTr), an effective Transformer-based backbone that can be applied in most voxel-based detectors to further enhance detection performance. 

\begin{figure}[!t] \centering   
\subfigure[3D convolutional network] {
 \label{intro_1}     
\includegraphics[width=0.225\textwidth]{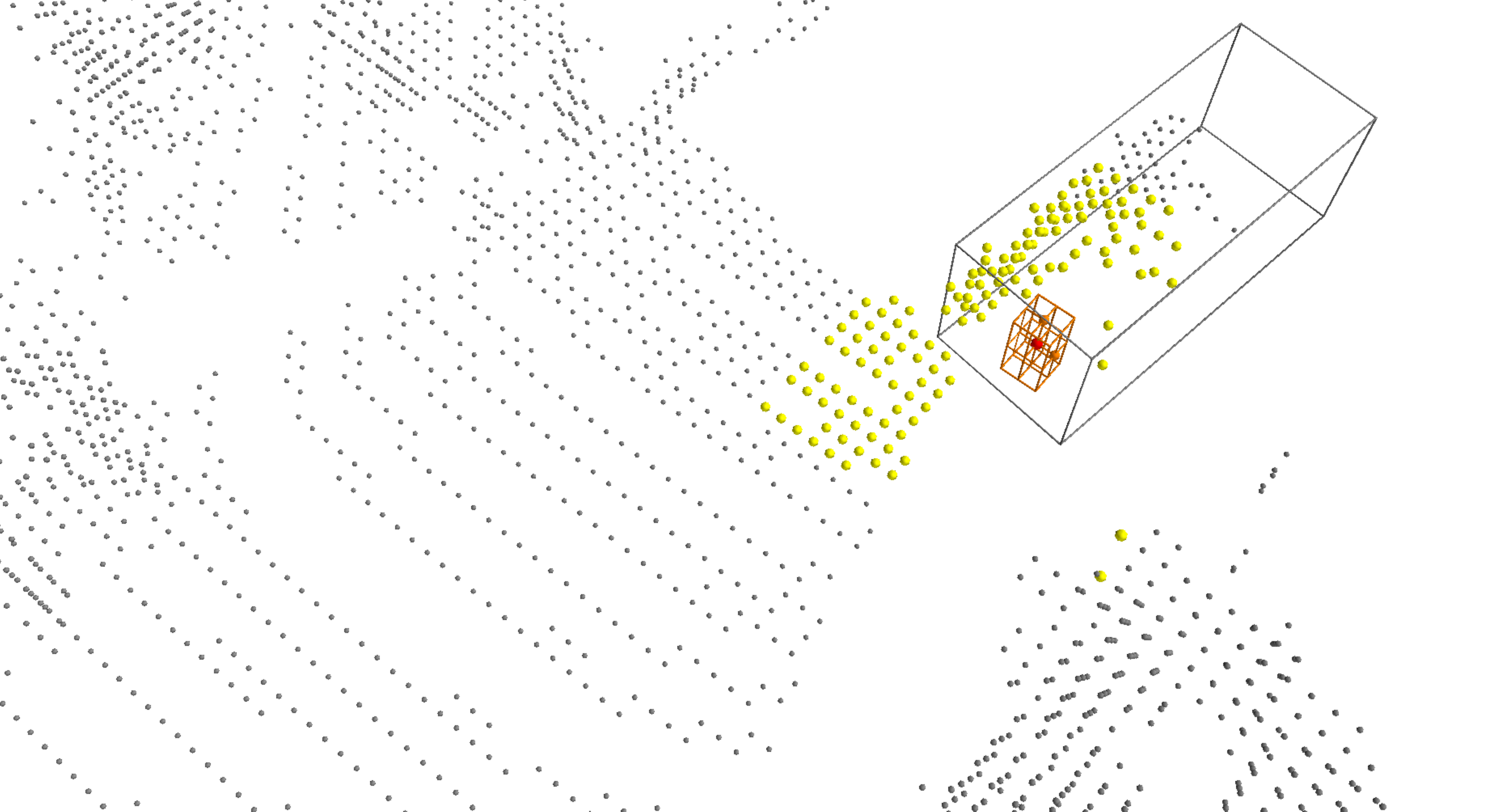}  
}    
\subfigure[Voxel Transformer] { 
\label{intro_2}     
\includegraphics[width=0.225\textwidth]{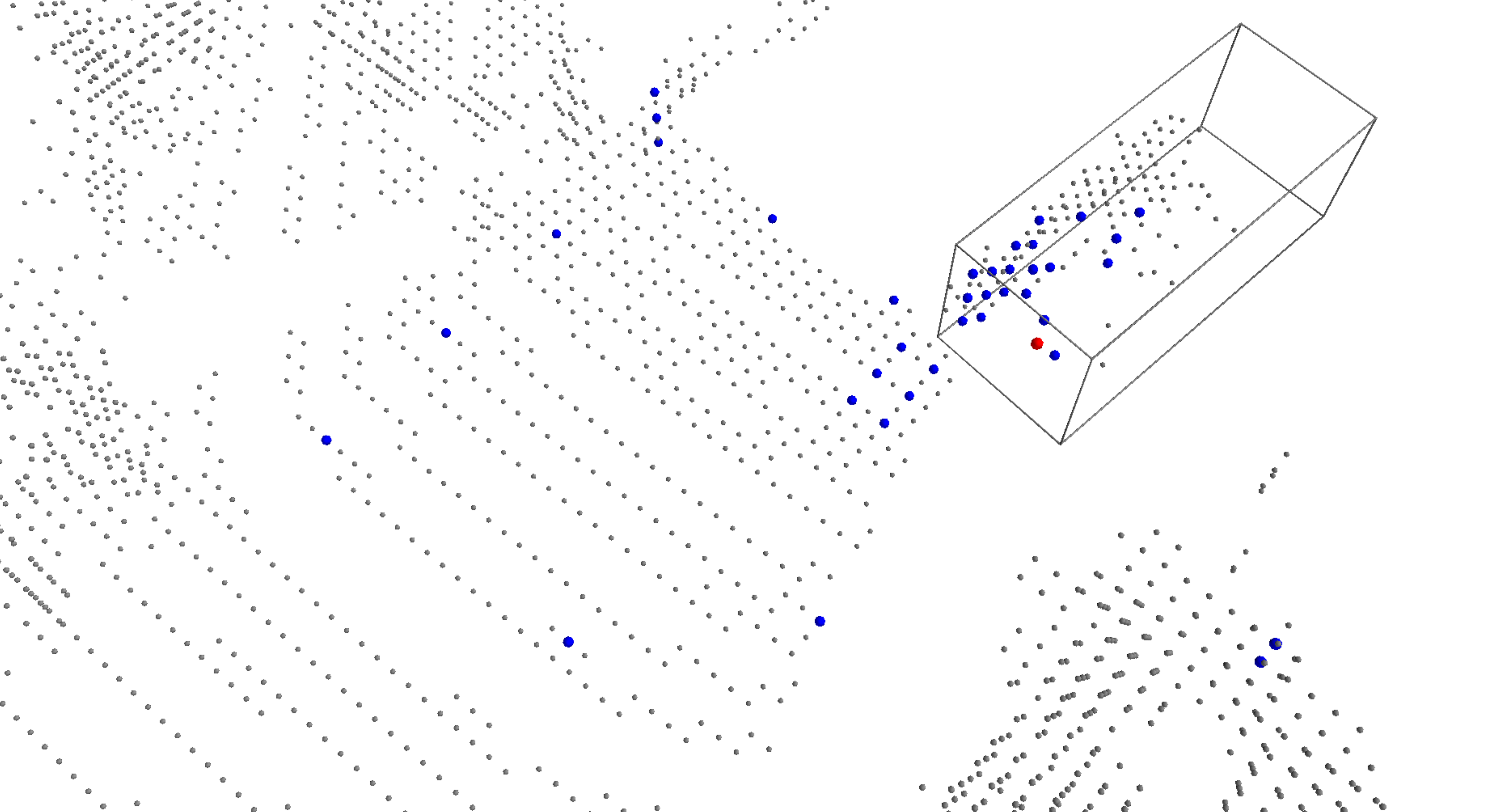}     
}
\caption{Illustration of the receptive field obtained by the 3D convolutional network and our proposed VoTr. In \subref{intro_1}, the orange cube denotes a single 3D convolutional kernel, and the yellow voxels are covered by the maximum receptive field centered at the red voxel. In \subref{intro_2}, the red voxel denotes a querying voxel, and the blue voxels are the respective attending voxels for this query in voxel attention. Our observation is that a single self-attention layer in VoTr can cover a larger region than the whole convolutional backbone, and it can also maintain enough fine-grained 3D structures.}     
\label{fig_intro} 
\vspace{-3mm}
\end{figure}

Previous approaches can be divided into two branches. Point-based approaches~\cite{shi2019pointrcnn, pan20203d, yang20203dssd, yang2019std} directly operate and generate 3D bounding boxes on point clouds. Those approaches generally apply point operators~\cite{qi2017pointnet, mao2019} to extract features directly from point clouds, but suffer from the sparse and non-uniform point distribution and the time-consuming process of sampling and searching for neighboring points. Alternatively, voxel-based approaches~\cite{zhou2018voxelnet, yan2018second, yin2020center, mao2021one, ye2020hvnet} first rasterize point clouds into voxels and apply 3D convolutional networks to extract voxel features, and then voxels are transformed into a Bird-Eye-View (BEV) feature map and 3D boxes are generated on the BEV map. Compared with the point-based methods which heavily rely on time-consuming point operators, voxel-based approaches are more efficient with sparse convolutions, and can achieve state-of-the-art detection performance.

The 3D sparse convolutional network is a crucial component in most voxel-based detection models. Despite its advantageous efficiency, the 3D convolutional backbones cannot capture rich context information with limited receptive fields, which hampers the detection of 3D objects that have only a few voxels. For instance, with a commonly-used 3D convolutional backbone~\cite{yan2018second} and the voxel size as $(0.05m, 0.05m, 0.1m)$ on the KITTI dataset, the maximum receptive field in the last layer is only $(3.65m, 3.65m, 7.3m)$, which can hardly cover a car with the length over $4m$. Enlarging the receptive fields is also intractable. The maximum theoretical receptive field of each voxel is roughly proportional to the product of the voxel size $V$, the kernel size $K$, the downsample stride $S$, and the layer number $L$. Enlarging $V$ will lead to the high quantization error of point clouds. Increasing $K$ leads to the cubic growth of convoluted features. Increasing $S$ will lead to a low-resolution BEV map which is detrimental to the box prediction, and increasing $L$ will add much computational overhead. Thus it is computationally extensive to obtain large receptive fields for the 3D convolutional backbones. Given the fact that the large receptive field is heavily needed in detecting 3D objects which are naturally sparse and incomplete, a new architecture should be designed to encode richer context information compared with the convolutional backbone.   

Recently advances~\cite{dosovitskiy2020image, carion2020end, zheng2020rethinking} in 2D object classification, detection, and segmentation show that Transformer is a more effective architecture compared with convolutional neural networks, mainly because long-range relationships between pixels can be built by self-attention in the Transformer modules. However, directly applying standard Transformer modules to voxels is infeasible, mainly owing to two facts: 1) Non-empty voxels are sparsely distributed in a voxel-grid. Different from pixels which are densely placed on an image plane, non-empty voxels only account for a small proportion of total voxels, \eg, the non-empty voxels normally occupy less than $0.1\%$ of the total voxel space on the Waymo Open dataset~\cite{sun2020scalability}. Thus instead of performing self-attention on the whole voxel-grids, special operations should be designed to only attend to those non-empty voxels efficiently. 2) The number of non-empty voxels is still large in a scene, \eg, there are nearly $90k$ non-empty voxels generated per frame on the Waymo Open dataset. Therefore applying fully-connected self-attention like the standard Transformer is computationally prohibitive. New methods are thus highly desired to enlarge the attention range while keeping the number of attending voxels for each query in a small value.   

To this end, we propose Voxel Transformer (VoTr), a Transformer-based 3D backbone that can be applied upon voxels efficiently and can serve as a better substitute for the conventional 3D convolutional backbones. To effectively handle the sparse characteristic of non-empty voxels, we propose the sparse voxel module and the submanifold voxel module as the basic building blocks of VoTr. The submanifold voxel modules operate strictly on the non-empty voxels, to retain the original 3D geometric structure, while the sparse voxel modules can output features at the empty locations, which is more flexible and can further enlarge the non-empty voxel space. To resolve the problem that non-empty voxels are too numerous for self-attention, we further propose two attention mechanisms: Local Attention and Dilated Attention, for multi-head attention in the sparse and submanifold voxel modules. Local Attention focuses on the neighboring region to preserve detailed information. Dilated Attention obtains a large attention range with only a few attending voxels, by gradually increasing the search step. To further accelerate the querying process for Local and Dilated Attention, we propose Fast Voxel Query, which contains a GPU-based hash table to efficiently store and lookup the non-empty voxels. Combining all the above components, VoTr significantly boosts the detection performance compared with the convolutional baselines, while maintains computational efficiency.

Our main contributions can be summarized as follows:\\
\indent 1) We propose Voxel Transformer, the first Transformer-based 3D backbone for voxel-based 3D detectors.\\
\indent 2) We propose the sparse and submanifold voxel module to handle the sparsity characteristic of voxels, and we further propose special attention mechanisms and Fast Voxel Query for efficient computation.\\ 
\indent 3) Our VoTr consistently outperforms the convolutional baselines and achieves the state-of-the-art performance with $74.95\%$ LEVEL\_1 mAP for vehicle and $82.09\%$ mAP for moderate car class on the Waymo dataset and the KITTI dataset respectively.

\begin{figure*}[!t]
\centering
\includegraphics[width=0.9\textwidth]{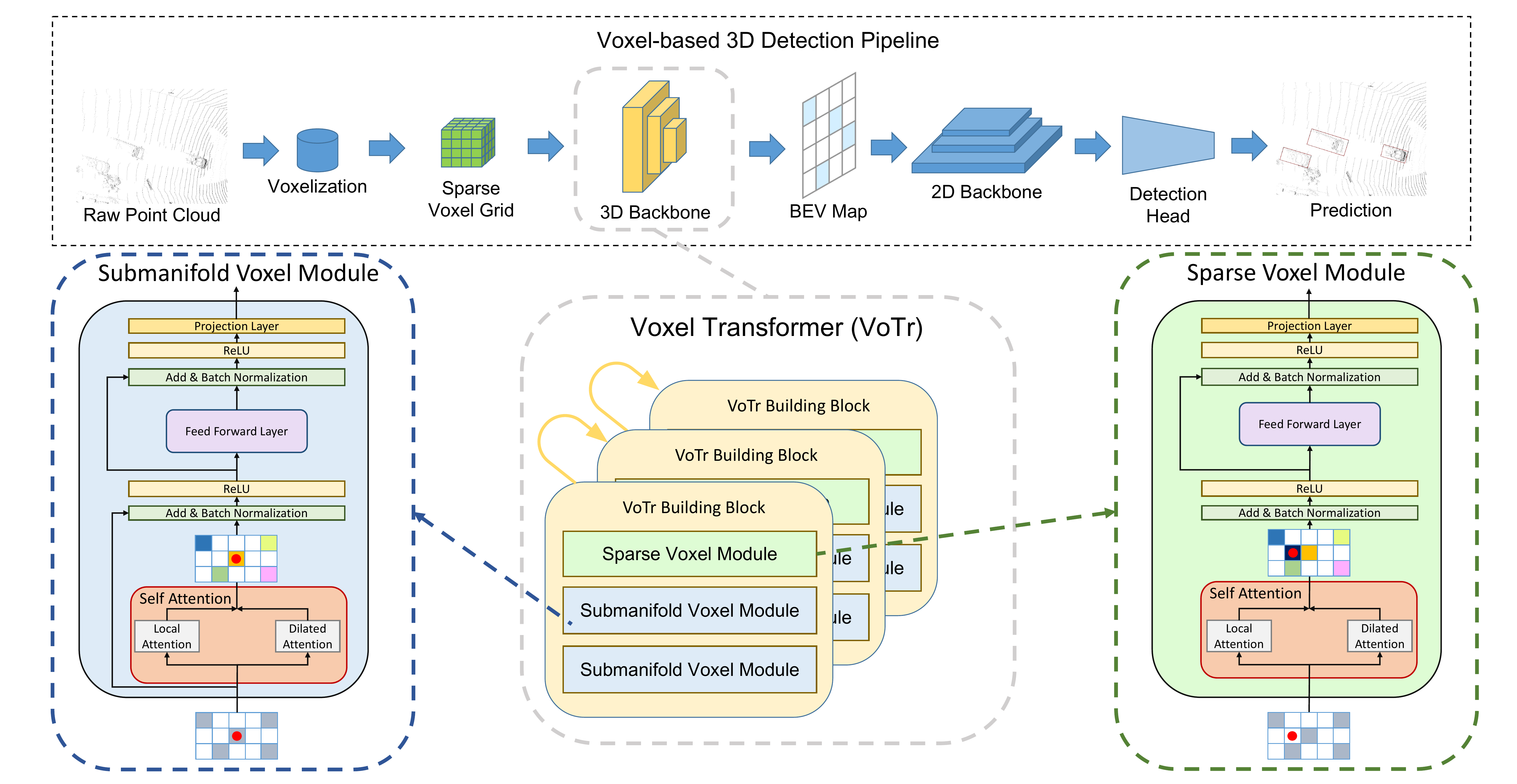}
\caption{The overall architecture of Voxel Transformer (VoTr). VoTr is a Transformer-based 3D backbone that can be applied in most voxel-based 3D detection frameworks. It contains a series of sparse and submanifold voxel modules. Submanifold voxel modules perform multi-head self-attention strictly on the non-empty voxels, while sparse voxel modules can extract voxel features at empty locations.}
\label{fig_framework}
\end{figure*}

\section{Related Work}
\noindent\textbf{3D object detection from point clouds.} 3D object detectors can be divided into $2$ streams: point-based and voxel-based. Point-based detectors operate directly on raw point clouds to generate 3D boxes. F-PointNet~\cite{qi2018frustum} is a pioneering work that utilizes frustums for proposal generation. PointRCNN~\cite{shi2019pointrcnn} generates 3D proposals from the foreground points in a bottom-up manner. 3DSSD~\cite{yang20203dssd} introduces a new sampling strategy for point clouds. Voxel-based detectors transform point clouds into regular voxel-grids and then apply 3D and 2D convolutional networks to generate 3D proposals. VoxelNet~\cite{zhou2018voxelnet} utilizes a 3D CNN to extract voxel features from a dense grid. SECOND~\cite{yan2018second} proposes 3D sparse convolutions to efficiently extract voxel features. HVNet~\cite{ye2020hvnet} designs a convolutional network that leverages the hybrid voxel representation. PV-RCNN~\cite{shi2020pv} uses keypoints to extract voxel features for boxes refinement. Point-based approaches suffer from the time-consuming process of sampling and aggregating features from irregular points, while voxel-based methods are more efficient owing to the regular structure of voxels. Our Voxel Transformer can be plugged into most voxel-based detectors to further enhance the detection performance while maintaining computational efficiency.

\noindent\textbf{Transformers in computer vision.} Transformer~\cite{vaswani2017attention} introduces a fully attentional framework for machine translation. Recently Transformer-based architectures surpass the convolutional architectures and show superior performance in the task of image classification, detection and segmentation. Vision Transformer~\cite{dosovitskiy2020image} splits an image into patches and feeds the patches into a Transformer for image classification. DETR~\cite{carion2020end} utilizes a Transformer-based backbone and a set-based loss for object detection. SETR~\cite{zheng2020rethinking} applies progressive upsampling on a Transformer-based backbone for semantic segmentation. MaX-DeepLab~\cite{wang2020max} utilizes a mask Transformer for panoptic segmentation. Transformer-based architectures are also used in 3D point clouds. Point Transformer~\cite{zhao2020point} designs a novel point operator for point cloud classification and segmentation. Pointformer~\cite{pan20203d} introduces attentional operators to extract point features for 3D object detection. Our Voxel Transformer extends the idea of Transformers on images, and proposes a novel method to apply Transformer to sparse voxels. Compared with point-based Transformers, Voxel Transformer benefits from the efficiency of regular voxel-grids and shows superior performance in 3D object detection. 

\section{Voxel Transformer}
In this section, we present Voxel Transformer (VoTr), a Transformer-based 3D backbone that can be applied in most voxel-based 3D detectors. VoTr can perform multi-head attention upon the empty and non-empty voxel positions though the sparse and submanifold voxel modules, and long-range relationships between voxels can be constructed by efficient attention mechanisms. We further propose Fast Voxel Query to accelerate the voxel querying process in multi-head attention. We will detail the design of each component in the following sections.   

\subsection{Overall Architecture}
In this section, we introduce the overall architecture of Voxel Transformer. Similar to the design of the conventional convolutional architecture~\cite{yan2018second} which contains $3$ sparse convolutional blocks and $6$ submanifold convolutional blocks, our VoTr is composed of a series of sparse and submanifold voxel modules, as shown in Figure~\ref{fig_framework}. In particular, we design $3$ sparse voxel modules which downsample the voxel-grids by $3$ times and output features at different voxel positions and resolutions as inputs. Each sparse voxel module is followed by $2$ submanifold voxel modules, which keeps the input and output non-empty locations the same, to maintain the original 3D structure while enlarge receptive fields. Multi-head attention is performed in all those modules, and the attending voxels for each querying voxel in multi-head attention are determined by two special attention mechanisms: Local Attention and Dilated Attention, which captures well diverse context in different ranges. Fast Voxel Query is further proposed to accelerate the searching process for the non-empty voxels in multi-head attention. 

Voxel features extracted by our proposed VoTr are then projected to a BEV feature map to generate 3D proposals, and the voxels and corresponding features can also be utilized on the second stage for RoI refinement. We note that our proposed VoTr is flexible and can be applied in most voxel-based detection frameworks~\cite{yan2018second, shi2020pv, deng2020voxel}.

\subsection{Voxel Transformer Module}
In this section, we present the design of sparse and submanifold voxel modules. The major difference between sparse and submanifold voxel modules is that submanifold voxel modules strictly operate on the non-empty voxels and extract features only at the non-empty locations, which maintains the geometric structures of 3D scenes, while sparse voxel modules can extract voxel features at the empty locations, which shows more flexibility and can expand the original non-empty voxel space according to needs. We first introduce self-attention on sparse voxels and then detail the design of sparse and submanifold voxel modules.

\textbf{Self-attention on sparse voxels.} We define a dense voxel-grid, which has $N_{dense}$ voxels in total, to rasterize the whole 3D scene. In practice we only maintain those non-empty voxels with a $N_{sparse} \times 3$ integer indices array $\mathcal{V}$ and $N_{sparse} \times d$ corresponding feature array $\mathcal{F}$ for efficient computation, where $N_{sparse}$ is the number of non-empty voxels and $N_{sparse} \ll N_{dense}$. In each sparse and submanifold voxel module, multi-head self-attention is utilized to build long-range relationships among non-empty voxels. Specifically, given a querying voxel $i$, the attention range $\Omega(i) \subseteq \mathcal{V}$ is first determined by attention mechanisms, and then we perform multi-head attention on the attending voxels $j \in \Omega(i)$ to obtain the feature $f^{attend}_{i}$. Let $f_{i}, f_{j} \in \mathcal{F}$ be the features of querying and attending voxels respectively, and $v_{i}, v_{j} \in \mathcal{V}$ be the integer indices of querying and attending voxels. We first transform the indices $v_{i}, v_{j}$ to the corresponding 3D coordinates of the real voxel centers $p_{i}, p_{j}$ by $p = r \cdot (v+0.5)$, where $r$ is the voxel size. Then for a single head, we compute the query embedding $Q_{i}$, key embedding $K_{j}$ and value embedding $V_{j}$ as:  
\begin{equation} \label{3.1.1}
    Q_{i} = f_{i}W_{q}, K_{j} = f_{j}W_{k} + E_{pos}, V_{j} = f_{j}W_{v} + E_{pos},
\end{equation}
where $W_{q}, W_{k}, W_{v}$ are the linear projection of query, key and value respectively, and the positional encoding $E_{pos}$ can be calculated by: 
\begin{equation} \label{3.1.2}
    E_{pos} = (p_{i} - p_{j})W_{pos}.
\end{equation}
Thus self-attention on voxels can be formulated as:
\begin{equation} \label{3.1.3}
    f_{i}^{attend} = \sum_{j \in \Omega(i)} \sigma(\frac{Q_{i}K_{j}}{\sqrt{d}}) \cdot V_{j},
\end{equation}
where $\sigma(\cdot)$ is the softmax normalization function. We note that self-attention on voxels is a natural 3D extension of standard 2D self-attention with sparse inputs and relative coordinates as positional embeddings.

\textbf{Submanifold voxel module.} The outputs of submanifold voxel modules are exactly at the same locations with the input non-empty voxels, which indicates its ability to keep the original 3D structures of inputs. In the submanifold voxel module, two sub-layers are designed to capture the long-range context information for each non-empty voxel. The first sub-layer is the self-attention layer that combines all the attention mechanisms, and the second is a simple feed-forward layer in~\cite{vaswani2017attention}. Residual connections are employed around the sub-layers. The major differences between the standard Transformer module and our proposed module are as three folds: 1) We append an additional linear projection layer after the feed-forward layer for channel adjustment of voxel features. 2) We replace layer normalization with batch normalization. 3) We remove all the dropout layers in the module, since the number of attending voxels is already small and randomly rejecting some of those voxels hampers the learning process.

\begin{figure}[!t]
\centering
\includegraphics[width=0.45\textwidth]{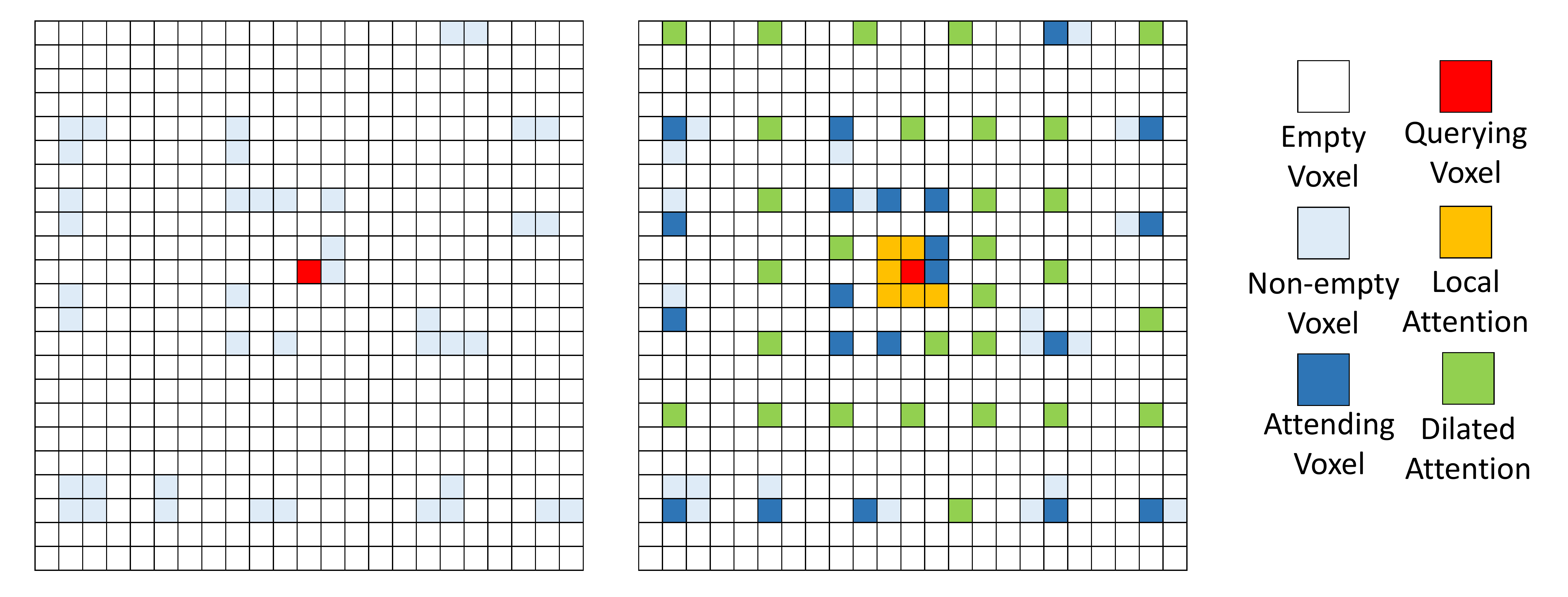}
\caption{Illustration of Local and Dilated Attention. We note that this is a 2D example and can be easily extended to 3D cases. For each query (red), Local Attention (yellow) focuses on the local region while Dilated Attention (green) searches the whole space with gradually enlarged steps. The non-empty voxels (light blue) which meet the searching locations are selected as the attending voxels (dark blue).}
\label{fig_attention}
\end{figure}

\textbf{Sparse voxel module.} Different from the submanifold voxel module which only operates on the non-empty voxels, the sparse voxel module can extract features for the empty locations, leading to the expansion of the original non-empty space, and it is typically required in the voxel downsampling process~\cite{yan2018second}. Since there is no feature $f_{i}$ available for the empty voxels, we cannot obtain the query embedding $Q_{i}$ from $f_{i}$. To resolve the problem, we give an approximation of $Q_{i}$ at the empty location from the attending features $f_{j}$:
\begin{equation} \label{3.1.4}
    Q_{i} = \mathop{\mathcal{A}}\limits_{j \in \Omega(i)}(f_{j}),
\end{equation}
where the function $\mathcal{A}$ can be interpolation, pooling, \etc. In this paper, we choose $\mathcal{A}$ as the maxpooling of all the attending features $f_{j}$. We also use Eq.\ref{3.1.3} to compute multi-head attention. The architecture of sparse voxel modules is similar to submanifold voxel modules, except that we remove the first residual connection around the self-attention layer, since the inputs and outputs are no longer the same.

\begin{figure*}[!t]
\centering
\includegraphics[width=0.9\textwidth]{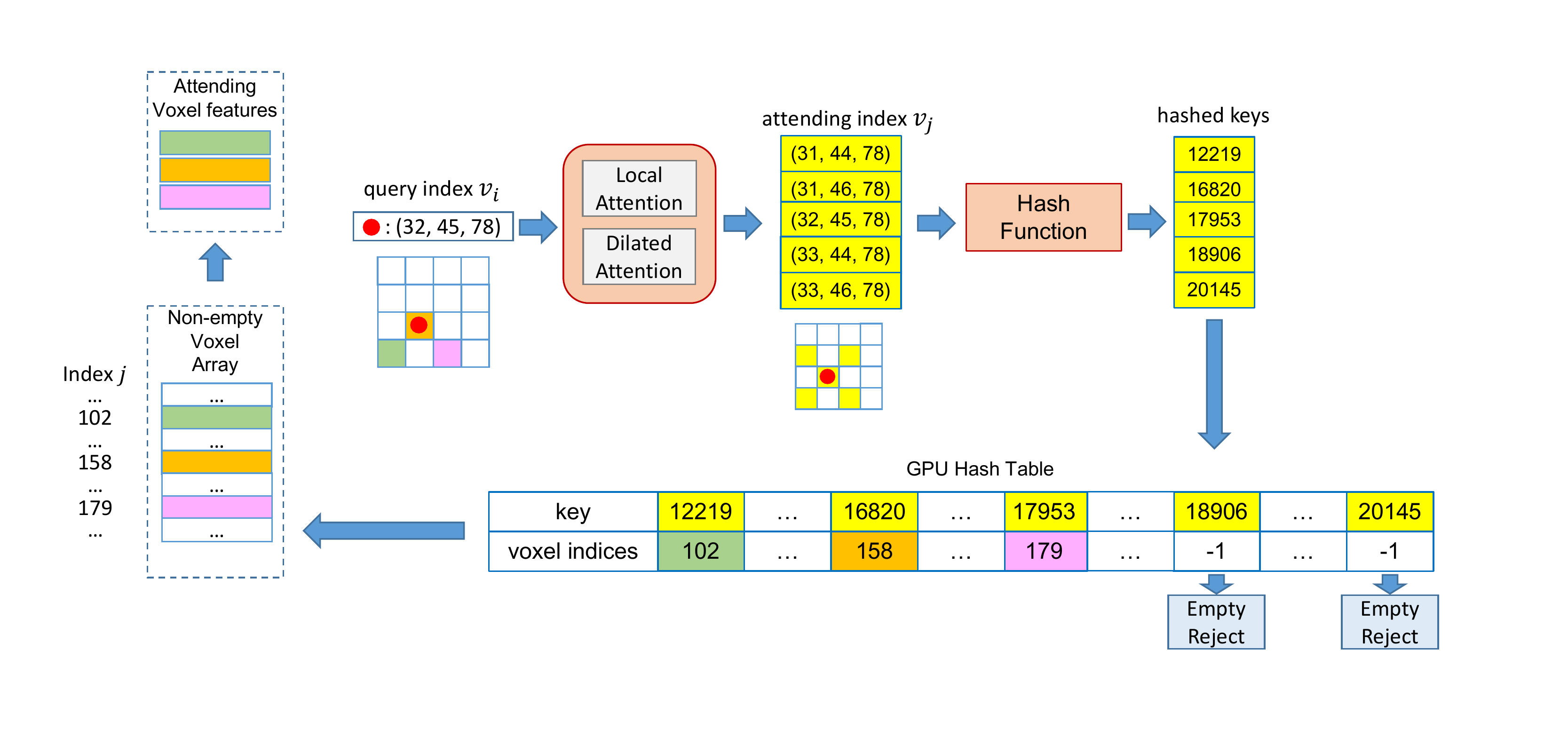}
\caption{Illustration of Fast Voxel Query. For each querying index $v_{i}$, an attending voxel index $v_{j}$ is determined by Local and Dilated Attention. Then we can lookup the non-empty index $j$ in the hash table with hashed $v_{j}$ as the key. Finally, the non-empty index $j$ is used to gather the attending feature $f_{j}$ from $\mathcal{F}$ for multi-head attention. Our proposed Fast Voxel Query is efficient both in time and in space and can significantly accelerate the computation of sparse voxel attention.}
\label{fig_fastvq}
\end{figure*}

\subsection{Efficient Attention Mechanism}
In this section, we delve into the design of the attention range $\Omega(i)$, which determines the attending voxels for each query $i$, and is a crucial factor in self-attention on sparse voxels. $\Omega(i)$ is supposed to satisfy the following requirements: 1) $\Omega(i)$ should cover the neighboring voxels to retain the fine-grained 3D structure. 2) $\Omega(i)$ should reach as far as possible to obtain a large context information. 3) the number of attending voxels in $\Omega(i)$ should be small enough, \eg less than $50$, to avoid heavy computational overhead. To tackle those issues, we take the inspiration from~\cite{zaheer2020big} and propose two attention mechanisms: Local Attention and Dilated Attention to control the attention range $\Omega(i)$. The designs of the two mechanisms are as follows.

\textbf{Local Attention.} We define $\varnothing(start, end, stride)$ as a function that returns the non-empty indices in a closed set $[start, end]$ with the step as $stride$. In the 3D cases, for example, $\varnothing((0,0,0), (1,1,1), (1,1,1))$ searches the set $\{(0,0,0), (0,0,1), (0,1,0), \cdots, (1,1,1)\}$ with $8$ indices for the non-empty indices. In Local Attention, given a querying voxel $v_{i}$, the local attention range $\Omega_{local}(i)$ parameterized by $R_{local}$ can be formulated as: 
\begin{equation} \label{3.1.5}
    \Omega_{local}(i) = \varnothing(v_{i} - R_{local}, v_{i} + R_{local}, (1,1,1)),
\end{equation}
where $R_{local} = (1,1,1)$ in our experiments. Local Attention fixes the $stride$ as $(1,1,1)$ to exploit every non-empty voxel inside the local range $R_{local}$, so that the fine-grained structures can be retained by Local Attention.

\textbf{Dilated Attention.} The attention range $\Omega_{dilated}(i)$ of Dilated Attention is defined by a parameter list $R_{dilated}$: $[(R^{(1)}_{start}, R^{(1)}_{end}, R^{(1)}_{stride}),\cdots,(R^{(M)}_{start}, R^{(M)}_{end}, R^{(M)}_{stride})]$, and the formulation of $\Omega_{dilated}(i)$ can be represented as:
\begin{equation} \label{3.1.6}
\begin{aligned}
    \Omega_{dilated}(i) =\bigcup^{M}_{m=1} & \varnothing(v_{i} - R^{(m)}_{end}, v_{i} + R^{(m)}_{end}, R^{(m)}_{stride}) \setminus \\& \varnothing(v_{i} - R^{(m)}_{start}, v_{i} + R^{(m)}_{start}, R^{(m)}_{stride}),
\end{aligned}
\end{equation}
where $\setminus$ is the set subtraction operator and the function $\bigcup$ takes the union of all the non-empty voxel sets. We note that $R^{(i)}_{start} < R^{(i)}_{end}\leq R^{(i+1)}_{start}$ and $R^{(i)}_{stride} < R^{(i+1)}_{stride}$, which means that we gradually enlarge the querying step $R^{(i)}_{stride}$ when search for the non-empty voxels which are more distant. This leads to a fact that we preserve more attending voxels near the query while still maintaining some attending voxels that are far away, and $R^{(i)}_{stride} > (1,1,1)$ significantly reduces the searching time and memory cost. With a carefully designed parameter list $R_{dilated}$, the attention range is able to reach more than $15m$ but the number of attending voxels for each querying voxel is still kept less than $50$. It is worth noting that Local Attention can be viewed as a special case in Dilated Attention when $R_{start}=(0,0,0)$, $R_{end}=(1,1,1)$ and $R_{stride}=(1,1,1)$.


\subsection{Fast Voxel Query}
Searching for the non-empty attending voxels for each query is non-trivial in voxel self-attention. The sparse indices array $\mathcal{V}$ cannot arrange 3D sparse voxel indices in order in one dimension $N_{sparse}$. Thus we cannot directly obtain the index $j \in \Omega(i)$ in $\mathcal{V}$,
even if we can easily get the corresponding integer voxel index $v_{j} \in \mathbb{R}^{3}$. Iterating all the $N_{sparse}$ non-empty voxels to find the matched $j$ takes $O(N_{sparse})$ time complexity for each querying process, and it is extremely time-consuming since $N_{sparse}$ is normally $90k$ on the Waymo Open dataset. In~\cite{deng2020voxel} dense 3D voxel-grids are utilized to store $j$ (or $-1$ if empty) for all the empty and non-empty voxels, but it is extremely memory-consuming to maintain those dense 3D voxel-grids, where the total number of voxels $N_{dense}$ is more than $10^{7}$. In this paper, we propose Fast Voxel Query, a new method that applies a GPU-based hash table to efficiently look up the attending non-empty voxels with little memory consumption.  

An illustration of Fast Voxel Query is shown in Figure~\ref{fig_fastvq}. Fast Voxel Query consists of four major steps: 1) we build a hash-table on GPUs which stores the hashed non-empty integer voxel indices $v_{j}$ as keys, and the corresponding indices $j$ for the array $\mathcal{V}$ as values. 2) For each query $i$, we apply Local Attention and Dilated Attention to obtain the attending voxel indices $v_{j} \in \Omega(i)$. 3) We look up the respective indices $j$ for $\mathcal{V}$ using the hashed key $v_{j}$ in the hash table, and $v_{j}$ is judged as an empty voxel and rejected if the hash value returns $-1$. 4) We can finally gather the attending voxel indices $v_{j}$ and features $f_{j}$ from $\mathcal{V}$ and $\mathcal{F}$ with $j$ for voxel self-attention. We note that all the steps can be conducted in parallel on GPUs by assigning each querying voxel $i$ a separate CUDA thread, and in the third step, the lookup process for each query only costs $O(N_{\Omega})$ time complexity, where $N_{\Omega}$ is the number of voxels in $\Omega(i)$ and $N_{\Omega} \ll N_{sparse}$.

To leverage the spatial locality of GPU memory, we build the hash table as a $N_{hash} \times 2$ tensor, where $N_{hash}$ is the hash table size and $N_{sparse} < N_{hash} \ll N_{dense}$. The first row of the $N_{hash} \times 2$ hash table stores the keys and the second row stores the values. We use the linear probing scheme to resolve the collisions in the hash table, and the atomic operations to avoid the data race among CUDA threads. Compared with the conventional methods~\cite{qi2017pointnet++, deng2020voxel}, our proposed Fast Voxel Query is efficient both in time and in space, and our approach remarkably accelerates the computation of voxel self-attention.

\section{Experiments}
\begin{table*}[!t]
\setlength{\tabcolsep}{4.5mm}{
\begin{tabular}{l|c|c|c c c}
\toprule
\multirow{2}{*}{Methods} & \multirow{1}{*}{LEVEL\_1} & \multirow{1}{*}{LEVEL\_2} & \multicolumn{3}{c}{LEVEL\_1 3D mAP/mAPH by Distance} \\
 &   3D mAP/mAPH           &       3D mAP/mAPH      
 & 0-30m            & 30-50m          & 50m-Inf         \\
\midrule
PointPillars~\cite{lang2019pointpillars}   & 63.3/62.7  & 55.2/54.7  & 84.9/84.4  & 59.2/58.6    & 35.8/35.2       \\
MVF~\cite{MVF}  & 62.93/-      & -      & 86.30/-     & 60.02/-         & 36.02/-         \\
Pillar-OD~\cite{wang2020pillar}     & 69.8/-   & -   & 88.5/-    & 66.5/-    & 42.9/-     \\
AFDet~\cite{ge2020afdet}  & 63.69/-    &  -  & 87.38/-  & 62.19/-    & 29.27/-         \\
LaserNet~\cite{meyer2019lasernet}  & 52.1/50.1  & -   & 70.9/68.7   & 52.9/51.4  & 29.6/28.6 \\
CVCNet~\cite{chen2020every}  & 65.2/-  & -   & 86.80/-    & 62.19/-    & 29.27/-         \\
StarNet~\cite{ngiam2019starnet} & 64.7/56.3   & 45.5/39.6 & 83.3/82.4& 58.8/53.2 & 34.3/25.7 \\
RCD~\cite{bewley2020range} & 69.0/68.5  & - & 87.2/86.8  & 66.5/66.1    & 44.5/44.0       \\
Voxel R-CNN~\cite{deng2020voxel}  & 75.59/-  & 66.59/-  & 92.49/-    & 74.09/-   & 53.15/-  \\
\midrule
SECOND$^{\star}$~\cite{yan2018second} & 67.94/67.28 & 59.46/58.88  & 88.10/87.46 & 65.31/64.61 & 40.36/39.57 \\
\textbf{VoTr-SSD (ours)}    &  \textbf{68.99/68.39}  &  \textbf{60.22/59.69}    & \textbf{88.18/87.62}     &  \textbf{66.73/66.05}   &   \textbf{42.08/41.38}   \\
\midrule
PV-RCNN~\cite{shi2020pv}   & 70.3/69.7                           & 65.4/64.8                           & 91.9/91.3        & 69.2/68.5       & 42.2/41.3       \\
PV-RCNN$^{\star}$ \cite{shi2020pv}                  & 74.06/73.38        & 64.99/64.38                           & -        & -       & -       \\
\textbf{VoTr-TSD (ours)}    & \textbf{74.95/74.25} & \textbf{65.91/65.29} 
     & \textbf{92.28/91.73}   & \textbf{73.36/72.56}    & \textbf{51.09/50.01}           \\
\bottomrule
\end{tabular}}
\setlength{\belowcaptionskip}{10pt}
\caption{Performance comparison on the Waymo Open Dataset with 202 validation sequences for the vehicle detection. $\star$: re-implemented with the official code.} \label{table_waymo_1}
\end{table*}

In this section, we evaluate Voxel Transformer on the commonly used Waymo Open dataset~\cite{sun2020scalability} and the KITTI~\cite{geiger2013vision} dataset. We first introduce the experimental settings and two frameworks based on VoTr, and then compare our approach with previous state-of-the-art methods on the Waymo Open dataset and the KITTI dataset. Finally, we conduct ablation studies to evaluate the effects of different configurations.

\subsection{Experimental Setup} \label{Experimental Setup}
\noindent\textbf{Waymo Open Dataset.} The Waymo Open Dataset contains $1000$ sequences in total, including $798$ sequences (around $158k$ point cloud samples) in the training set and 202 sequences (around $40k$ point cloud samples) in the validation set. The official evaluation metrics are standard 3D mean Average Precision (mAP) and mAP weighted by heading accuracy (mAPH). Both of the two metrics are based on an IoU threshold of 0.7 for vehicles and 0.5 for other categories. The testing samples are split in two ways. The first way is based on the distances of objects to the sensor: $0-30m$, $30-50m$ and $>50m$. The second way is according to the difficulty levels: LEVEL\_1 for boxes with more than five LiDAR points and LEVEL\_2 for boxes with at least one LiDAR point.

\noindent\textbf{KITTI Dataset.} The KITTI dataset contains $7481$ training samples and $7518$ test samples, and the training samples are further divided into the \textit{train} split ($3712$ samples) and the $val$ split ($3769$ samples). The official evaluation metric is mean Average Precision (mAP) with a rotated IoU threshold 0.7 for cars. On the \textit{test} set mAP is calculated with $40$ recall positions by the official server. The results on the \textit{val} set are calculated with 11 recall positions for a fair comparison with other approaches.

We provide $2$ architectures based on Voxel Transformer: VoTr-SSD is a single-stage voxel-based detector with VoTr as the backbone. VoTr-TSD is a two-stage voxel-based detector based on VoTr.

\noindent\textbf{VoTr-SSD.} \textit{\textbf{Vo}xel \textbf{Tr}ansformer for \textbf{S}ingle-\textbf{S}tage \textbf{D}etector} is built on the commonly-used single-stage framework SECOND~\cite{yan2018second}. In particular, we replace the 3D sparse convolutional backbone of SECOND, with our proposed Voxel Transformer as the new backbone, and we still use the anchor-based assignment following~\cite{yan2018second}. Other modules and configurations are kept the same for a fair comparison.

\noindent\textbf{VoTr-TSD.} \textit{\textbf{Vo}xel \textbf{Tr}ansformer for \textbf{T}wo-\textbf{S}tage \textbf{D}etector} is built upon the state-of-the-art two-stage framework PV-RCNN~\cite{shi2020pv}. Specifically, we replace the 3D convolutional backbone on the first stage of PV-RCNN, with our proposed Voxel Transformer as the new backbone, and we use keypoints to extract voxel features from Voxel Transformer for the second stage RoI refinement. Other modules and configurations are kept the same for a fair comparison.

\noindent\textbf{Implementation Details.} VoTr-SSD and VoTr-TSD share the same architecture on the KITTI and Waymo dataset. The input non-empty voxel coordinates are first transformed into $16$-channel initial features by a linear projection layer, and then the initial features are fed into VoTr for voxel feature extraction. The channels of voxel features are lifted up to $32$ and $64$ in the first and second sparse voxel module respectively, and other modules keep the input and output channels the same. Thus the final output features have $64$ channels. The number of total attending voxels is set to $48$ for each querying voxel, and the number of heads is set to $4$ for multi-head attention. The GPU hash table size $N_{hash}$ is set to $400k$. We would like readers to refer to supplementary materials for the detailed design of attention mechanisms.

\noindent\textbf{Training and Inference Details.} Voxel Transformer is trained along with the whole framework with the ADAM optimizer. On the KITTI dataset, VoTr-SSD and VoTr-TSD are trained with the batch size $32$ and $16$ respectively, and with the learning rate $0.01$ for $80$ epochs on $8$ V100 GPUs. On the Waymo Open dataset, we uniformly sample $20\%$ frames for training and use the full validation set for evaluation following~\cite{shi2020pv}. VoTr-SSD and VoTr-TSD are trained with the batch size $16$ and the learning rate $0.003$ for $60$ and $80$ epochs respectively on $8$ V100 GPUs. The cosine annealing strategy is adopted for the learning rate decay. Data augmentations and other configurations are kept the same as the corresponding baselines~\cite{yan2018second, shi2020pv}.

\begin{table}[]
\setlength{\tabcolsep}{2.1mm}{
\begin{tabular}{|l|c| c c c|}
\toprule
\multirow{2}{*}{Methods} & \multirow{2}{*}{Modality} & \multicolumn{3}{c}{$AP_{3D}$ (\%)} \\
                        &                           & Easy   & Mod.   & Hard   \\
\midrule
MV3D~\cite{mv3d}        & R+L                       & 74.97  & 63.63  & 54.00  \\
AVOD-FPN~\cite{avod}    & R+L                       & 83.07  & 71.76  & 65.73  \\
F-PointNet~\cite{fp}    & R+L                       & 82.19  & 69.79  & 60.59  \\
MMF~\cite{mmf}          & R+L                       & 88.40  & 77.43  & 70.22  \\
3D-CVF~\cite{3dcvf}     & R+L                       & 89.20  & 80.05  & 73.11  \\
CLOCs~\cite{clocs}      & R+L                       & 88.94  & 80.67  & 77.15  \\
ContFuse~\cite{cont}    & R+L                       & 83.68  & 68.78  & 61.67  \\
\midrule
VoxelNet~\cite{zhou2018voxelnet}    & L             & 77.47  & 65.11  & 57.73  \\
PointPillars~\cite{lang2019pointpillars}    & L     & 82.58  & 74.31  & 68.99  \\
PointRCNN~\cite{shi2019pointrcnn}   & L             & 86.96  & 75.64  & 70.70  \\
Part-$A^{2}$ Net~\cite{shi2020points} & L           & 87.81  & 78.49  & 73.51  \\
STD~\cite{yang2019std}          & L                 & 87.95  & 79.71  & 75.09  \\
Patches~\cite{patch}    & L                         & 88.67  & 77.20  & 71.82  \\
3DSSD~\cite{yang20203dssd}      & L                 & 88.36  & 79.57  & 74.55  \\
SA-SSD~\cite{he2020structure}   & L                 & 88.75  & 79.79  & 74.16  \\
TANet~\cite{tanet}      & L                         & 85.94  & 75.76  & 68.32  \\
Voxel R-CNN~\cite{deng2020voxel}    & L             & 90.90  & 81.62  & 77.06  \\
HVNet~\cite{ye2020hvnet}    & L                     & 87.21  & 77.58  & 71.79  \\
PointGNN~\cite{pointgnn}    & L                     & 88.33  & 79.47  & 72.29  \\
\midrule
SECOND~\cite{yan2018second}     & L                 & 84.65  & 75.96  & 68.71  \\
\textbf{VoTr-SSD (ours)}    & L         & \textbf{86.73} & \textbf{78.25} & \textbf{72.99}  \\
\midrule
PV-RCNN~\cite{shi2020pv}        & L                 & 90.25  & 81.43  & 76.82  \\
\textbf{VoTr-TSD (ours)}      & L & \textbf{89.90}  & \textbf{82.09}  & \textbf{79.14} \\ 
\bottomrule
\end{tabular}}
\setlength{\belowcaptionskip}{10pt}
\caption{Performance comparison on the KITTI \textit{test} set with AP calculated by $40$ recall positions for the car category. R+L denotes the methods that combines RGB data and point clouds. L denotes LiDAR-only approaches.} \label{table_kitti_1}
\vspace{-5mm}
\end{table}

\begin{table}[]
\setlength{\tabcolsep}{4.25mm}{
\begin{tabular}{|l| c c c|}
\toprule
\multirow{2}{*}{Methods} & \multicolumn{3}{c}{$AP_{3D}$ (\%)} \\
                                        & Easy   & Mod.   & Hard   \\
\midrule
PointRCNN~\cite{shi2019pointrcnn}       & 88.88  & 78.63  & 77.38  \\
STD~\cite{yang2019std}                  & 89.70  & 79.80  & 79.30  \\
3DSSD~\cite{yang20203dssd}              & 89.71  & 79.45  & 78.67  \\
VoxelNet~\cite{zhou2018voxelnet}        & 81.97  & 65.46  & 62.85  \\
Voxel R-CNN~\cite{deng2020voxel}        & 89.41  & 84.52  & 78.93  \\
PointPillars~\cite{lang2019pointpillars}& 86.62  & 76.06  & 68.91  \\
Part-$A^{2}$ Net~\cite{shi2020points}   & 89.47  & 79.47  & 78.54  \\
TANet~\cite{tanet}                      & 87.52  & 76.64  & 73.86  \\
SA-SSD~\cite{he2020structure}           & 90.15  & 79.91  & 78.78  \\
\midrule
SECOND~\cite{yan2018second}             & 87.43  & 76.48  & 69.10  \\
\textbf{VoTr-SSD (ours)}               & \textbf{87.86} & \textbf{78.27}  & \textbf{76.93} \\
\midrule
PV-RCNN~\cite{shi2020pv}                & 89.35  & 83.69  & 78.70  \\
\textbf{VoTr-TSD (ours)}              & \textbf{89.04} & \textbf{84.04}  & \textbf{78.68} \\
\bottomrule
\end{tabular}}
\setlength{\belowcaptionskip}{10pt}
\caption{Performance comparison on the KITTI \textit{val} split with AP calculated by $11$ recall positions for the car category.} \label{table_kitti_2}
\vspace{-4mm}
\end{table}

\subsection{Comparisons on the Waymo Open Dataset} \label{3D Detection on the Waymo Open Dataset} 
We conduct experiments on the Waymo Open dataset to verify the effectiveness of our proposed VoTr. As is shown in Table~\ref{table_waymo_1}, simply switching from the 3D convolutional backbone to VoTr gives $1.05\%$ and $3.26\%$ LEVEL\_1 mAP improvements for SECOND~\cite{yan2018second} and PV-RCNN~\cite{shi2020pv} respectively. In the range of 30-50m and 50m-Inf, VoTr-SSD gives $1.42\%$ and $1.72\%$ improvements, and VoTr-TSD gives $3.37\%$ and $4.83\%$ improvements on LEVEL\_1 mAP. The significant performance gains in the far away area show the importance of large context information obtained by VoTr to 3D object detection.

\subsection{Comparisons on the KITTI Dataset} \label{3D Detection on the KITTI Dataset} 
We conduct experiments on the KITTI dataset to validate the efficacy of VoTr. As is shown in the Table~\ref{table_kitti_1}, VoTr-SSD and VoTr-TSD brings $2.29\%$ mAP and $0.66\%$ mAP improvement on the moderate car class on the KITTI \textit{val} split. For the hard car class, VoTr-TSD achieves 79.14$\%$ mAP, outperforming all the previous approaches by a large margin, which indicates the long-range relationships between voxels captured by VoTr is significant for detecting 3D objects that only have a few points. The results on the \textit{val} split in Table~\ref{table_kitti_2} show that VoTr-SSD and VoTr-TSD outperform the baseline methods by $1.79\%$ and $0.35\%$ mAP for the moderate car class. Observations on the KITTI dataset are consistent with those on the Waymo Open dataset.

\subsection{Ablation Studies} \label{Ablation Studies}
\textbf{Effects of Local and Dilated Attention.} Table~\ref{attention_ablation_table} indicates that Dilated Attention guarantees larger receptive fields for each voxel and brings $2.79\%$ moderate mAP gain compared to using only Local Attention.

\textbf{Effects of dropout in Voxel Transformer.} Table~\ref{table_ablation_2} details the influence of different dropout rates to VoTr. We found that adding dropout layers in each module is detrimental to the detection performance. The mAP drops by $8.52\%$ with the dropout probability as $0.3$. 

\textbf{Effects of the number of attending voxels.} Table~\ref{table_ablation_3} shows that increasing the number of attending voxels from $24$ to $48$ boosts the performance by $1.19\%$, which indicates that a  voxel can obtain richer context information by involving more attending voxels in multi-head attention. 

\textbf{Comparisons on the model parameters.} Table~\ref{table_ablation_4} shows that replacing the 3D convolutional backbone with VoTr reduces the model parameters by $0.5M$, mainly because the modules in VoTr only contain linear projection layers, which have only a few parameters, while 3D convolutional kernels typically contain a large number of parameters. 

\textbf{Comparisons on the inference speed.} Table~\ref{table_ablation_5} shows that with carefully designed attention mechanisms and Fast Voxel Query, VoTr maintains computation efficiency with $14.65$ Hz running speed for the single-stage detector. Replacing the convolutional backbone with VoTr only adds about $20$ ms latency per frame. 

\begin{figure}[!t]
\centering
\includegraphics[width=0.45\textwidth]{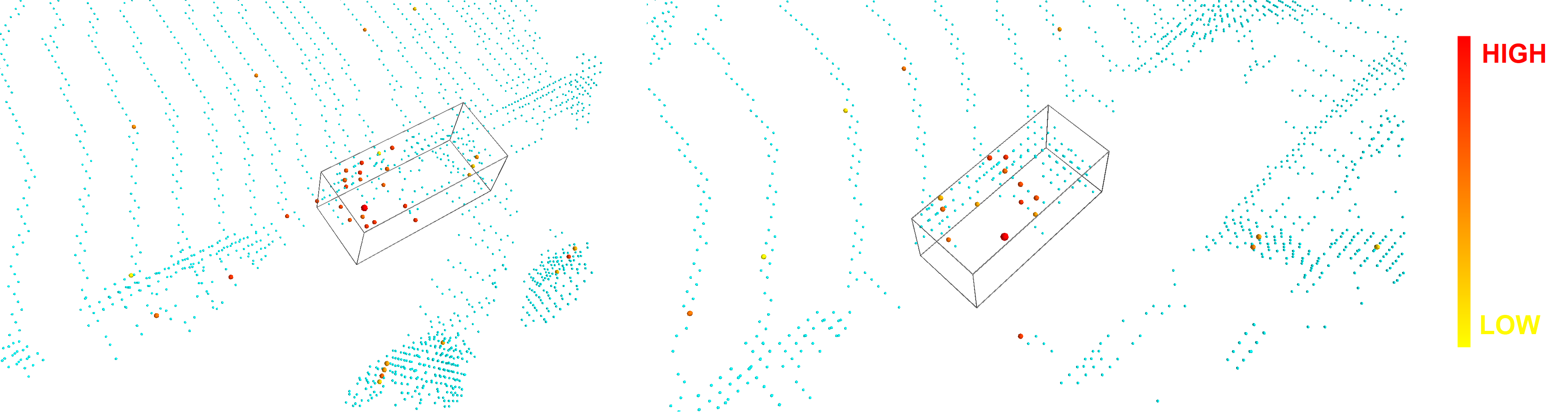}
\caption{Visualization of attention weights for attending voxels.}
\label{fig_weights}
\vspace{-2mm}
\end{figure}

\textbf{Visualization of attention weights.} Figure~\ref{fig_weights} shows that a querying voxel can dynamically select the features of attending voxels in a very large context range, which benefits the detection of objects that are sparse and incomplete.


\begin{table}[]
\setlength{\tabcolsep}{5.35mm}{
\begin{tabular}{|c|c c|c|}
\toprule
Methods  & L.A. & D.A. & AP$_{3D}$ ($\%$) \\
\midrule
(a)      & $\surd$    &            &  75.48         \\
(b)      & $\surd$    & $\surd$    &  \textbf{78.27}         \\
\bottomrule
\end{tabular}}
\setlength{\belowcaptionskip}{10pt}
\caption{Effects of attention mechanisms on the KITTI val split. L.A.: Local Attention. D.A.: Dilated Attention.}
\vspace{-2mm}
\label{attention_ablation_table}
\end{table}

\begin{table}[]
\setlength{\tabcolsep}{4.65mm}{
\begin{tabular}{|c|c|c|c|}
\toprule
Methods  &  Dropout probability  & AP$_{3D}$ ($\%$)  \\
\midrule
(a)      & $\textbf{0}$        & \textbf{78.27}         \\
(b)      & $0.1$      & 75.97         \\
(c)      & $0.2$      & 70.82         \\
(d)      & $0.3$      & 69.75         \\ 
\bottomrule
\end{tabular}}
\setlength{\belowcaptionskip}{10pt}
\caption{Effects of dropout probabilities on the KITTI \textit{val} split.} \label{table_ablation_2}
\vspace{-2mm}
\end{table}

\begin{table}[]
\setlength{\tabcolsep}{2.8mm}{
\begin{tabular}{|c|c|c|c|}
\toprule
Methods  &  Number of attending voxels  & AP$_{3D}$ ($\%$)  \\
\midrule
(a)      & $24$      & 77.08         \\
(b)      & $32$      & 77.72         \\
(c)      & $\textbf{48}$      & \textbf{78.27}        \\ 
\bottomrule
\end{tabular}}
\setlength{\belowcaptionskip}{10pt}
\caption{Effects of the number of attending voxels for each querying voxel on the KITTI \textit{val} split.} \label{table_ablation_3}
\vspace{-2mm}
\end{table}

\begin{table}[]
\setlength{\tabcolsep}{8mm}{
\begin{tabular}{|l|c|c|c|}
\toprule
Methods  &  Model parameters \\
\midrule
SECOND~\cite{yan2018second}      & $5.3M$      \\
\textbf{VoTr-SSD (ours)}     & $\textbf{4.8M}$      \\
\midrule
PV-RCNN~\cite{shi2020pv}      & $13.1M$      \\
\textbf{VoTr-TSD (ours)}     & $\textbf{12.6M}$ \\
\bottomrule
\end{tabular}}
\setlength{\belowcaptionskip}{10pt}
\caption{Comparisons on the model parameters for different frameworks on the KITTI dataset.} \label{table_ablation_4}
\vspace{-2mm}
\end{table}

\begin{table}[]
\setlength{\tabcolsep}{7mm}{
\begin{tabular}{|l|c|c|c|}
\toprule
Methods  &  Inference speed (Hz) \\
\midrule
SECOND~\cite{yan2018second}      & $20.73$      \\
\textbf{VoTr-SSD (ours)}     & $14.65$      \\
\midrule
PV-RCNN~\cite{shi2020pv}      & $9.25$      \\
\textbf{VoTr-TSD (ours)}     & $7.17$ \\
\bottomrule
\end{tabular}}
\setlength{\belowcaptionskip}{10pt}
\caption{Comparisons on the inference speeds for different frameworks on the KITTI dataset. 48 attending voxels are used.} \label{table_ablation_5}
\vspace{-2mm}
\end{table}

\section{Conclusion}
We present Voxel Transformer, a general Transformer-based 3D backbone that can be applied in most voxel-based 3D detectors. VoTr consists of a series of sparse and submanifold voxel modules, and can perform self-attention on sparse voxels efficiently with special attention mechanisms and Fast Voxel Query. For future work, we plan to explore more Transformer-based architectures on 3D detection.

{\small
\bibliographystyle{ieee_fullname}
\bibliography{egbib}
}
\clearpage
\appendix

\section{Architecture}
\begin{figure}[!t]
\centering
\includegraphics[width=0.45\textwidth]{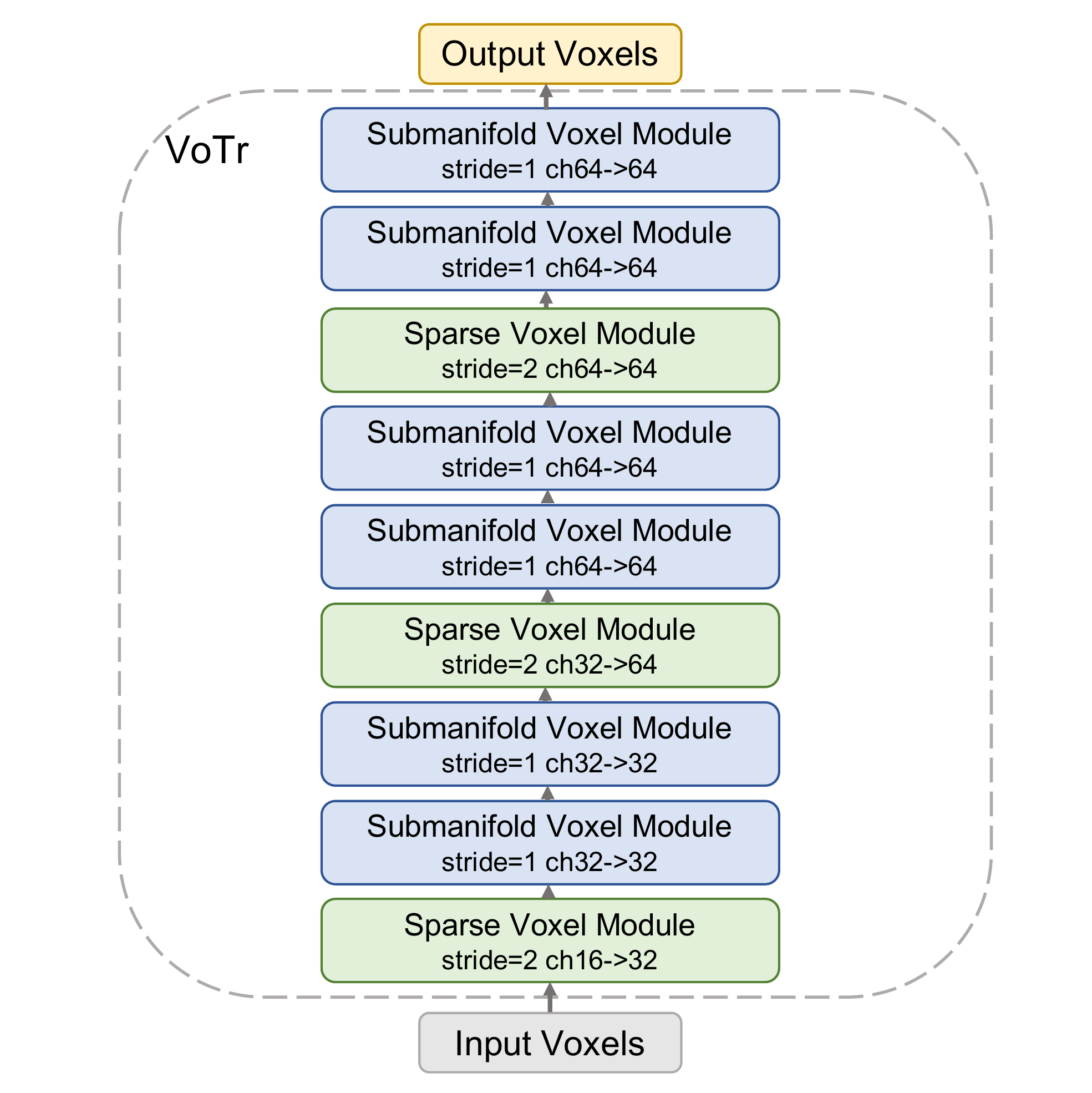}
\caption{Architecture of Voxel Transformer.}
\label{fig_achitect}
\end{figure}

\begin{figure}[!t]
\centering
\includegraphics[width=0.45\textwidth]{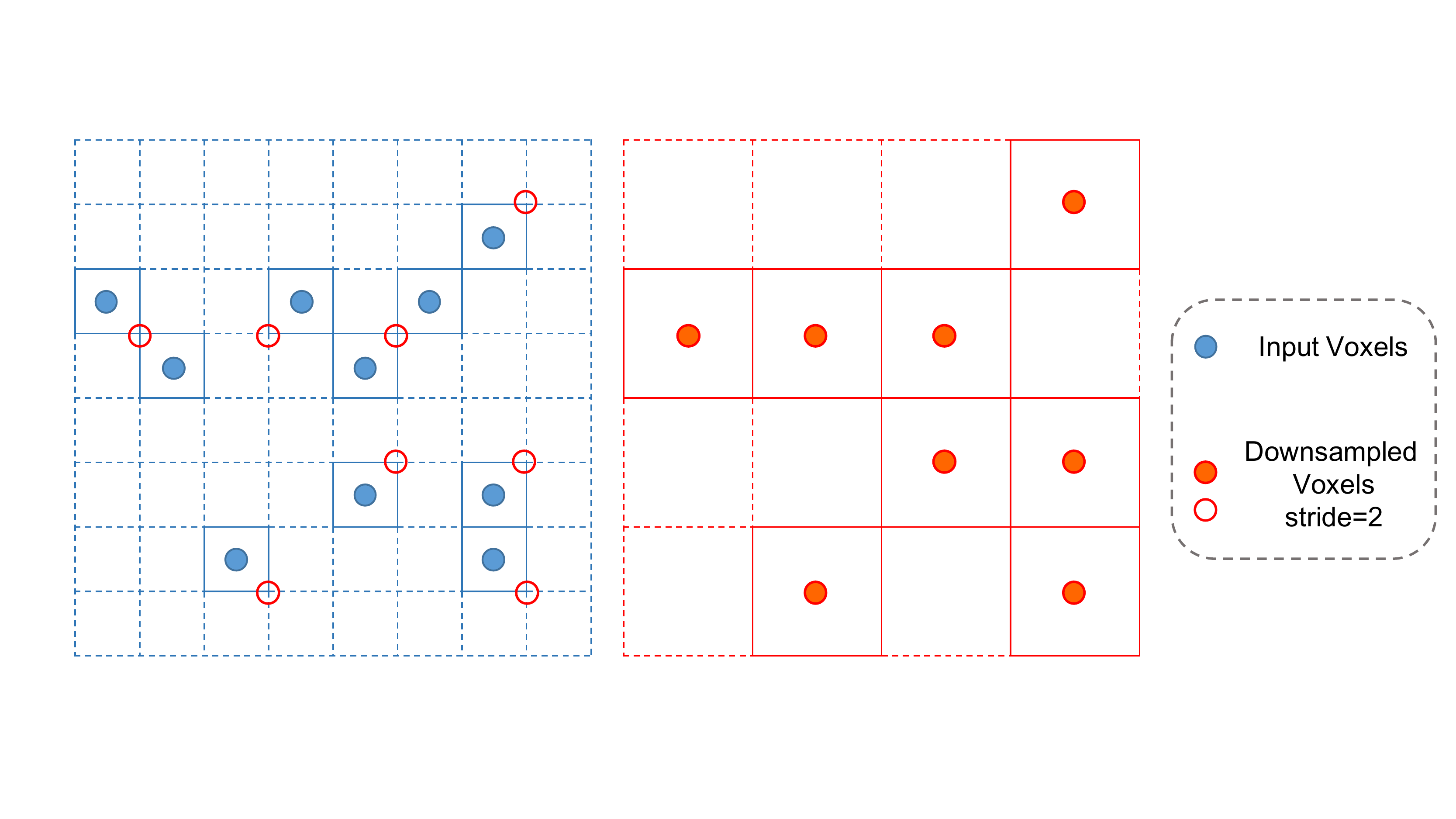}
\caption{Illustration of voxel downsampling process with stride $2$.}
\label{fig_downsample}
\end{figure}

The detailed architecture of Voxel Transformer is shown in Figure~\ref{fig_achitect}. Input voxels are downsampled $3$ times with the stride $2$ by $3$ sparse voxel modules. Figure~\ref{fig_downsample} shows an illustration of the voxel downsampling process. We note that the downsampled voxel centers are no longer overlapped with the original voxel centers, since the voxel size are doubled during downsampling. Thus sparse voxel modules are needed to perform voxel attention on those empty locations.  

\section{Dilated Attention}
In this section, we provide the configurations of Dilated Attention in Table~\ref{table_1}. We use the same configurations for both VoTr-TSD and VoTr-SSD on the KITTI dataset. With carefully designed Dilated Attention, a single self-attention layer can obtain large context information with only a few attending voxels.  

\begin{table}[!t]
\setlength{\tabcolsep}{3.55mm}{
\begin{tabular}{|c|c|c|c|}
\toprule
Module                          & $R_{start}$  & $R_{end}$   & $R_{stride}$ \\
\midrule
\multirow{3}{*}{1}      & (2,2,0)   & (5,5,3)      & (1,1,1)   \\
                               & (5,5,0)   & (25,25,15)   & (5,5,2)   \\
                               & (25,25,0) & (125,125,15) & (25,25,3) \\
\midrule
\multirow{3}{*}{2-4} & (2,2,0)   & (4,4,3)      & (1,1,1)   \\
                               & (4,4,0)   & (12,12,8)    & (3,3,2)   \\
                               & (12,12,0) & (60,60,8)    & (12,12,2) \\
\midrule
\multirow{3}{*}{5-7} & (2,2,0)   & (3,3,2)      & (1,1,1)   \\
                               & (3,3,0)   & (8,8,4)      & (2,2,1)   \\
                               & (8,8,0)   & (32,32,4)    & (8,8,1)   \\
\midrule
\multirow{2}{*}{8-9}      & (2,2,0)   & (4,4,3)      & (1,1,1)   \\
                               & (4,4,0)   & (16,16,5)    & (2,2,1)   \\
\bottomrule
\end{tabular}}
\setlength{\belowcaptionskip}{10pt}
\caption{Configurations of Dilated Attention on the KITTI dataset. The modules are indexed in a sequential order.} \label{table_1}
\end{table}

\section{Qualitative Results}
\begin{figure*}[!t] \centering   
\subfigure[] {
 \label{viz_1_1}     
\includegraphics[width=0.4\textwidth]{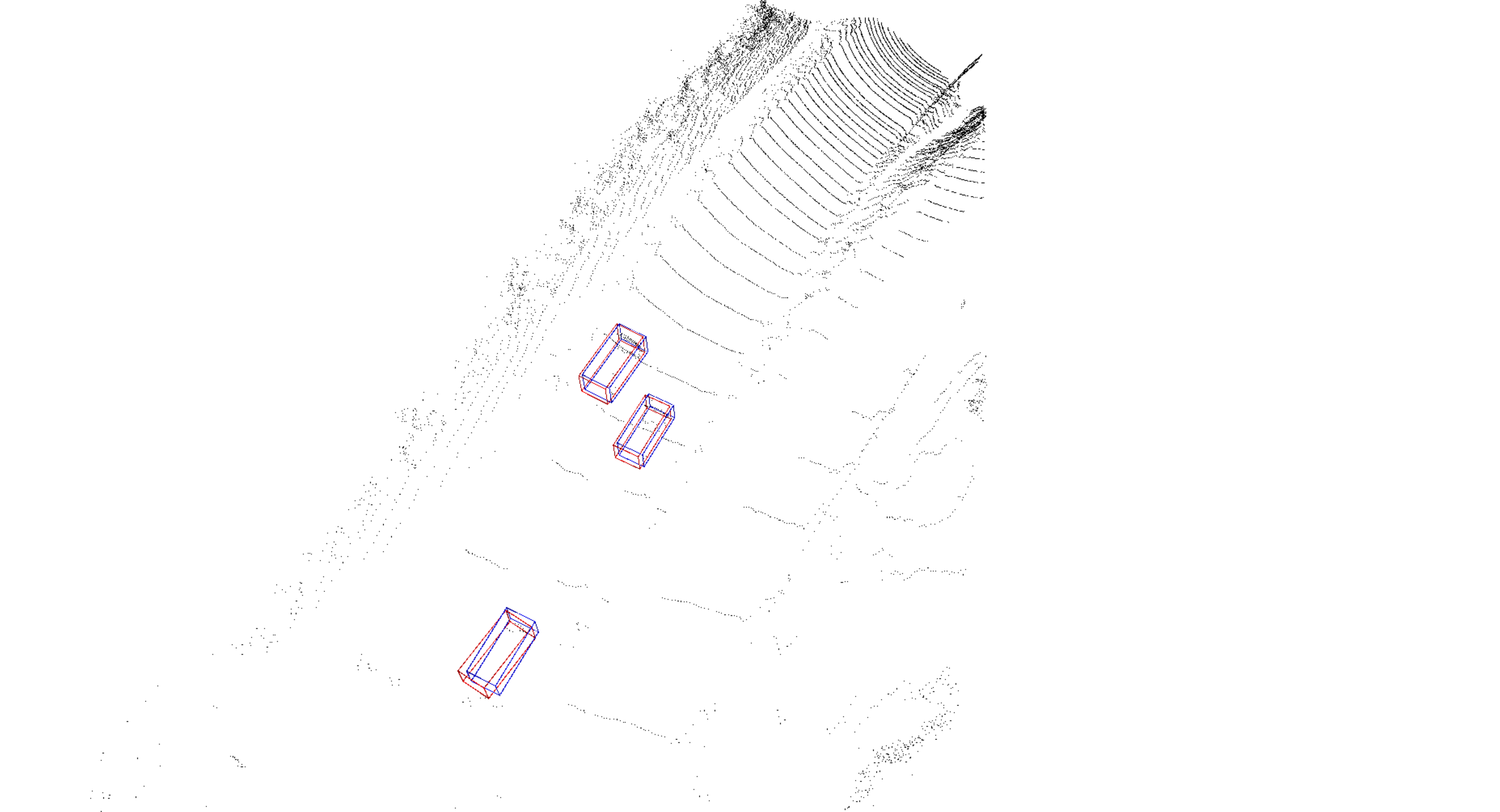}  
}    
\subfigure[] { 
\label{viz_1_2}     
\includegraphics[width=0.4\textwidth]{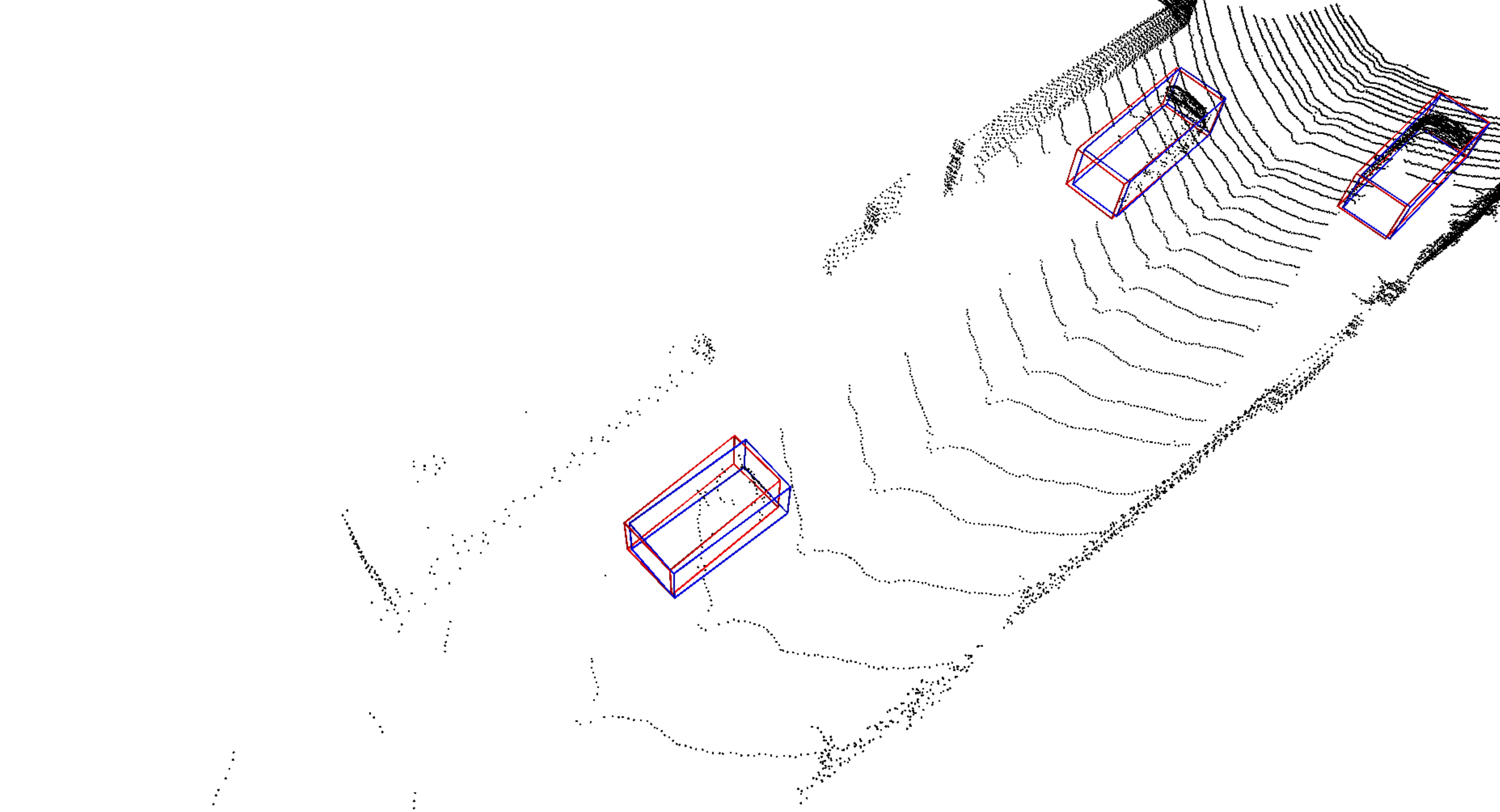}     
}
\subfigure[] { 
\label{viz_1_3}     
\includegraphics[width=0.4\textwidth]{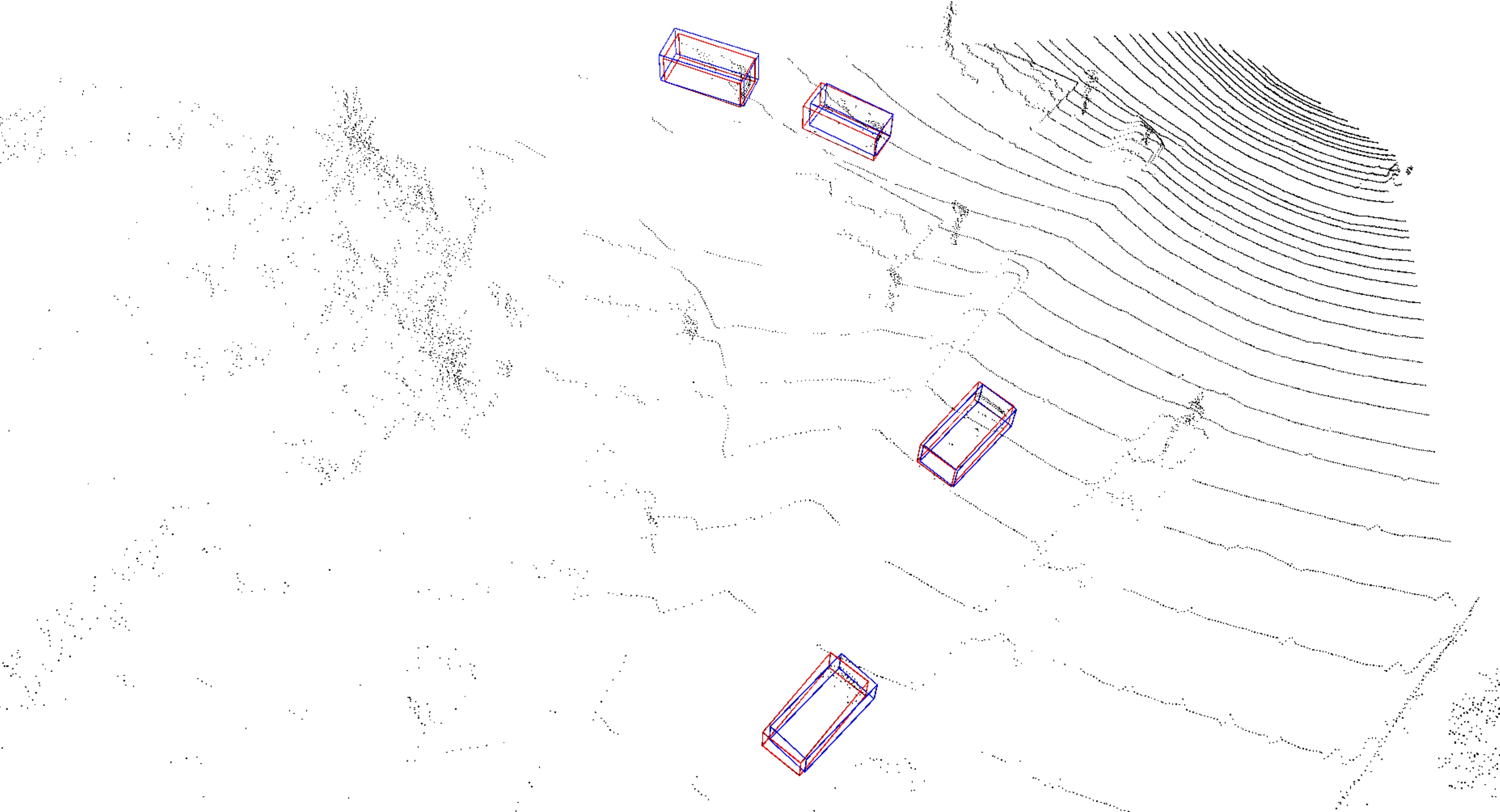}     
}
\caption{Visualization of detection results on the KITTI dataset. Blue boxes are the ground truth boxes, and red boxes are the boxes predicted by VoTr-TSD.}     
\label{fig_viz_1} 
\end{figure*}

\begin{figure*}[!t] \centering   
\subfigure[] {
 \label{viz_2_1}     
\includegraphics[width=0.4\textwidth]{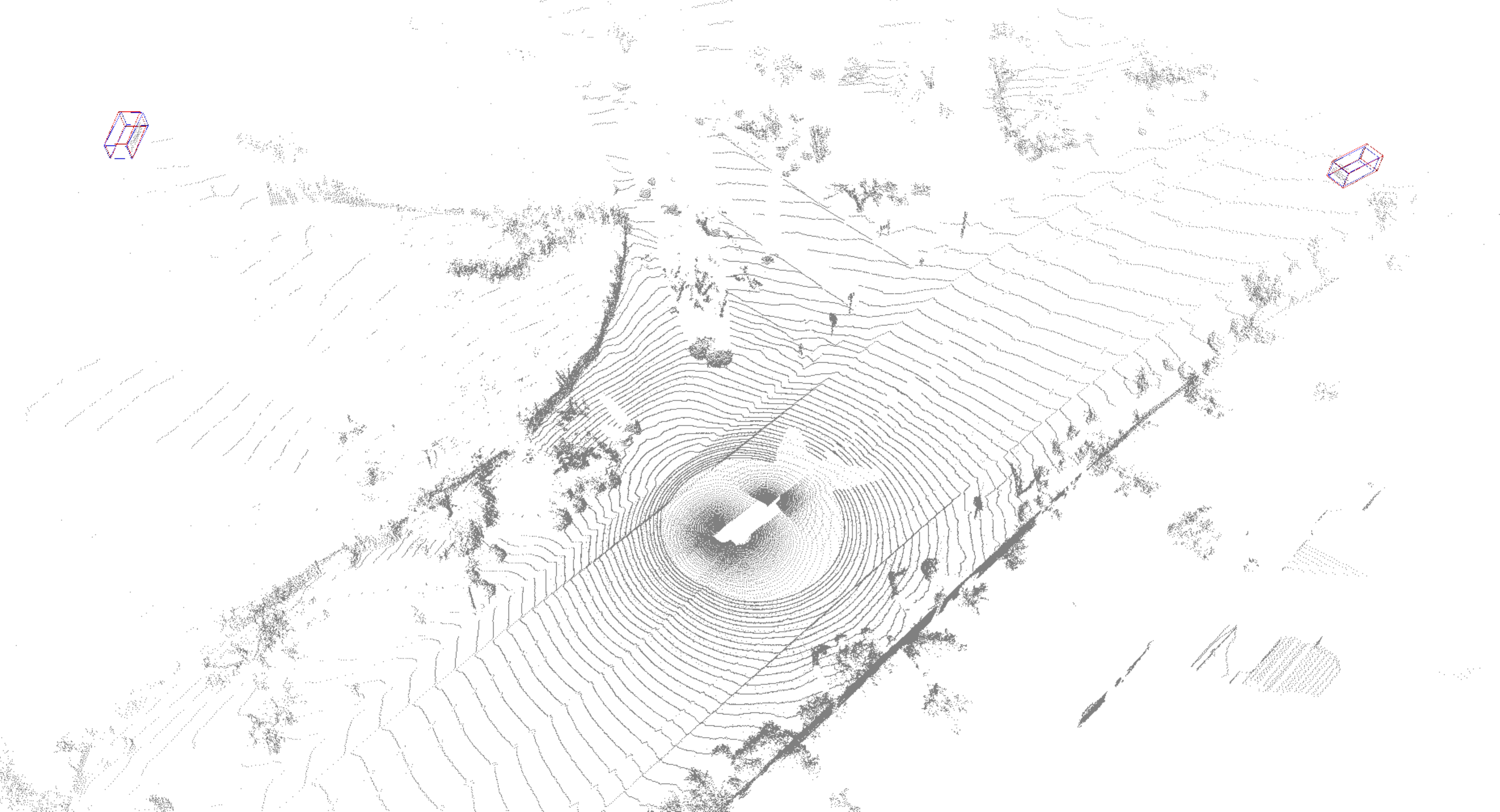}  
}    
\subfigure[] { 
\label{viz_2_2}     
\includegraphics[width=0.4\textwidth]{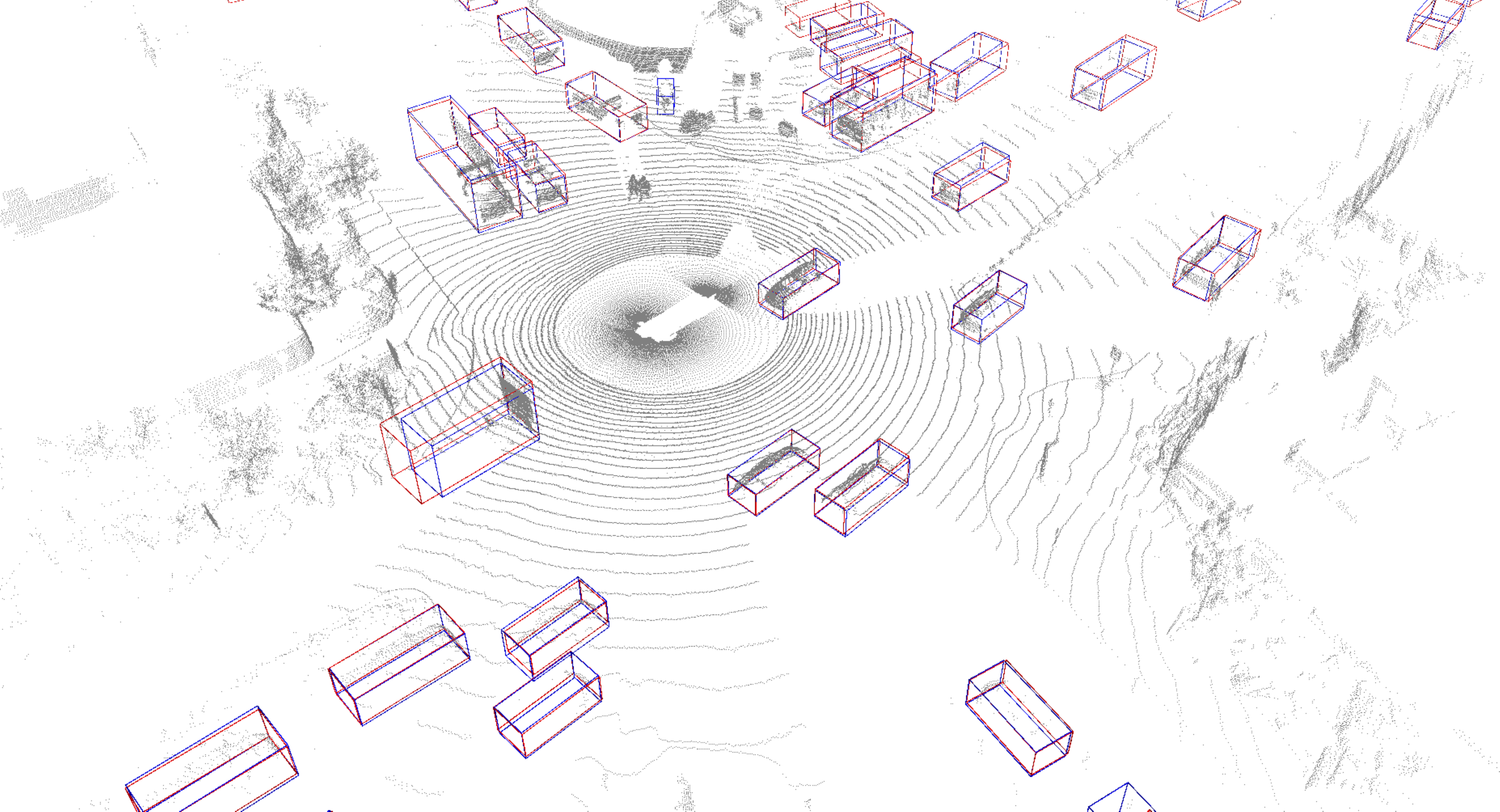}     
}
\subfigure[] { 
\label{viz_2_3}     
\includegraphics[width=0.4\textwidth]{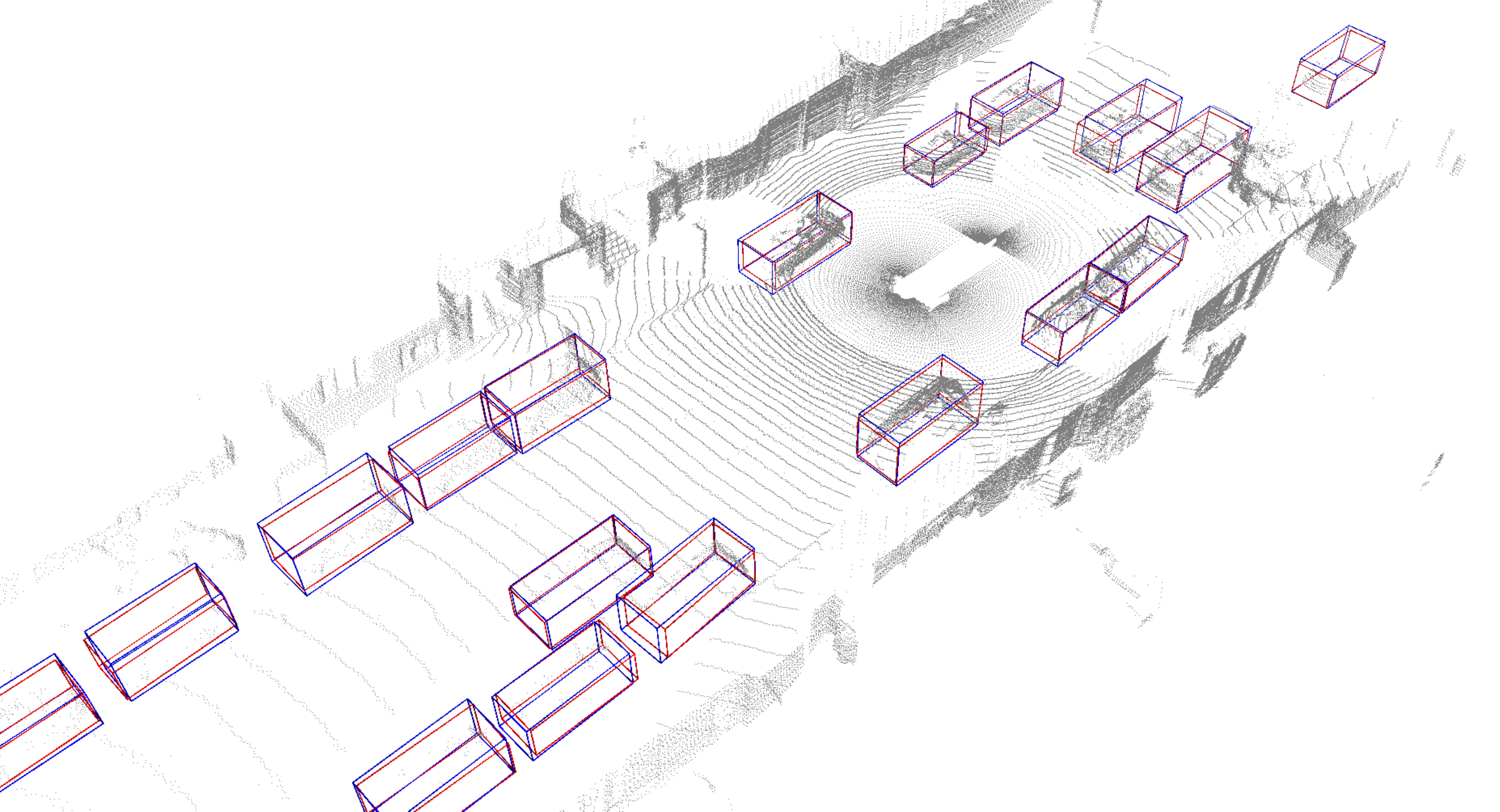}     
}
\caption{Visualization of detection results on the Waymo Open dataset. Blue boxes are the ground truth boxes, and red boxes are the boxes predicted by VoTr-TSD.}     
\label{fig_viz_2} 
\end{figure*}

In this section, we provide the qualitative results on the KITTI dataset in Figure~\ref{fig_viz_1}, and the Waymo Open dataset in Figure~\ref{fig_viz_2}. With rich context information captured by self-attention, our Voxel Transformer is able to detect those 3D objects that are sparse and incomplete effectively. 

\end{document}